\documentclass[conference]{IEEEtran}
\IEEEoverridecommandlockouts
\usepackage{cite}

\usepackage{amsmath,amssymb,amsfonts}
\usepackage[ruled,vlined]{algorithm2e}
\usepackage{algorithmic}
\usepackage{graphicx}
\usepackage{textcomp}
\usepackage{xcolor}
\usepackage{dblfloatfix}
\usepackage{subcaption}

\begin{document}

\title{Range Limited Coverage Control using Air-Ground Multi-Robot Teams}
\author{\IEEEauthorblockN{Max Rudolph}
\textit{Georgia Institute of Technology}\\
Atlanta, US \\
maxrudolph@gatech.edu

\and
\IEEEauthorblockN{Sean Wilson}
\textit{Georgia Institute of Technology}\\
Atlanta, US \\
sean.wilson@gtri.gatech.edu
\and
\IEEEauthorblockN{Magnus Egerstedt}
\textit{Georgia Institute of Technology}\\
Atlanta, US \\
magnus.egerstedt@ece.gatech.edu
}

\maketitle

\begin{abstract}

In this paper, we investigate how heterogeneous multi-robot systems with different sensing capabilities can observe a domain with an \emph{apriori} unknown density function. Common coverage control techniques are targeted towards homogeneous teams of robots and do not consider what happens when the sensing capabilities of the robots are vastly different. This work proposes an extension to Lloyd's algorithm that fuses coverage information from heterogeneous robots with differing sensing capabilities to effectively observe a domain. Namely, we study a bimodal team of robots consisting of aerial and ground agents. In our problem formulation we use aerial robots with coarse domain sensors to approximate the number of ground robots needed within their sensing region to effectively cover it. This information is relayed to ground robots, who perform an extension to the Lloyd's algorithm that balances a locally focused coverage controller with a globally focused distribution controller. The stability of the Lloyd's algorithm extension is proven and its performance is evaluated through simulation and experiments using the \emph{Robotarium}, a remotely-accessible, multi-robot testbed. 

\end{abstract}

\begin{IEEEkeywords}
Multi-Robot Systems, Distributed Robot Systems, Networked Robots, Distributed Sensor Networks
\end{IEEEkeywords}

\section{Introduction}

Multi-robot systems are well suited to solve highly parallelizable and redundant tasks \cite{guatam}, \cite{cao}, \cite{ota}. One such task is environmental monitoring and surveillance, where features or events within a large domain must be simultaneously observed. Environmental monitoring is commonly solved with a method called coverage control where agents in a multi-agent system distribute themselves throughout the domain optimally with respect to the features of interest \cite{Pimenta2}. Coverage control has many practical applications including distribution of farming robots and search and rescue teams surveying a disaster site \cite{lowenberg},\cite{zhong}. A family of coverage control solutions derived from Lloyd's algorithm presented in \cite{cortes} perform coverage control by having the agents follow the spatial gradient of a coverage quality cost function. In this paper, we address the implementation of coverage control on a team of heterogeneous agents consisting of aerial and ground robots with different but limited sensing capabilities that further extends Lloyd's algorithm.

Lloyd's algorithm has been extended upon in many ways in an effort to increase the algorithm’s applicability to partially heterogeneous and range-limited teams \cite{arslan}, \cite{hexsel}, \cite{cortesRangeLimited}. Previous extensions of Lloyd’s algorithm have introduced a weighting coefficient to the coverage cost function when used on heterogeneous teams \cite{santos}. The method proposed in \cite{arslan} uses power diagrams to account for the robots' differing capabilities but considers robots of slightly different sensing ranges. Research on coverage control in non-convex domains has shown that Lloyd’s algorithm can be extended to limited sensing robots operating on domain with a uniform density function without major loss of performance when compared to the traditional Lloyd's algorithm  \cite{rus2}, \cite{kantaros}. A practical variant of Lloyd's algorithm presented in \cite{cortesRangeLimited} uses proximity graphs to account for an agents’ limited sensing and communication capability in a domain with a non-uniform density. However, these extensions to Lloyd’s algorithm do not account for a method to fuse knowledge from significantly heterogeneous teams of robots.

In the coverage control problem formulation, agents operate over a domain with events or features to be observed that are distributed according to some underlying density function \cite{cortesVariations}. On a domain with a uniform density function, the cost function and subsequent distributed control law is determined only by the boundaries of the domain and the location of the agent’s Voronoi graph neighbours \cite{pimenta}. On a domain where the underlying density function varies spatially, the coverage cost function is formulated using the domain boundaries, agent neighbors, and the density function. Because of this, the density function must be known \emph{apriori}, agents must have large sensing ranges, or the domain must be explored to learn it. This poses a question when implementing Lloyd's algorithm on a system of robots that are surveying a domain—namely, how can robots effectively explore and cover a domain without complete knowledge of the underlying density function? This paper proposes a method to fuse information from a heterogeneous team of aerial and ground robots with different sensing capabilities to improve the coverage capabilities of the multi-robot system.

In this paper we consider a heterogeneous multi-robot team consisting of ground robots equipped with high resolution, low range sensors and aerial robots with low resolution, high range sensors, tasked with covering a domain much larger than an individual ground robots sensing range.  To overcome the limited global vision of the ground robots, the aerial robots roughly estimate the importance function within the domain and regions of interest throughout the domain are identified. This information can then be used by the ground robots to help distribute themselves globally while performing local coverage using their higher fidelity but range limited sensors. Combining the sensing capabilities of the aerial and ground robots within the system enables high resolution sensor coverage that would not be possible by either homogeneous team alone.  

With our proposed heterogeneous system, a trivial solution can be formed by simply using the aerial team to relay the entire density function (coarsely) to the ground team; however, this solution is not ideal for the following reasons; 
\begin{enumerate}
\item The fine sensing abilities of the first team of robots are not leveraged to make fine local positioning adjustments.
\vspace{1pt}
\item Relaying the entire density function can impose cumbersome, unnecessary, and potentially bandwidth restricted data transfers and computation when the multi-agent system becomes large.
\end{enumerate}
\noindent The method proposed in this work leverages the strengths within the heterogeneous team's sensing abilities and encodes critical global distribution data into a small matrix of region weights. Our method is a parameter-less extension to Lloyd’s algorithm that can leverage heterogeneous teams of robots to observe domains of non-uniform, multi-modal, unknown densities. 

Section II of this paper will describe the heterogeneous system and propose a coverage controller for the heterogeneous team with a proof of its stability. Section III will present results on the controller's performance in simulation and experimentally on the Robotarium \cite{robotarium}. Finally, Section IV will contain conclusions and discussions of the results.

\section{Coverage Control with Heterogeneous Teams}
\subsection{Traditional Lloyd's Algorithm}
In the traditional Lloyd's Algorithm, $M$ robots with position $p_i \subset \mathcal{D} \in \mathbb{R}^2, i = 1\dots M$ attempt to cover a convex domain $\mathcal{D}$. Each robot $i$ is given a region of dominance defined by a Voronoi cell
$$\mathcal{V}(p_i) = \{q \in \mathcal{D}\,|\, ||q - p_i|| \leq ||q - p_j||, \forall j \neq i\}.$$

\noindent With the domain split up into regions of sensing dominance, a cost function evaluating the quality of coverage with sensor quality decreasing with the inverse square of distance can be formulated as the following,

\begin{equation} \label{cost} \mathcal{H}(p) = \sum\limits_{i=1}^M\int_{\mathcal{V}(p_i)} ||q - p_i||^2 \phi(q) dq,\end{equation}

\noindent where $\phi$ is the underlying density function of the domain describing relative areas of interest such that $\phi : \mathcal{D} \rightarrow [0, \infty)$. As is shown in \cite{rus}, the following controller will drive a  robot team with infinite sensing capabilities to asymptotically achieve a centroidal Voronoi tessellation necessary for optimal coverage with respect to a stationary minimum of equation \ref{cost}, 

$$ u_i = \frac{\partial \mathcal{H}}{\partial p_i} \text{ and } \dot p_i = \kappa (c(p_i)-p_i)$$ 

\noindent where $\kappa > 0$ is a control gain and $c(p_i)$ is the mass center of the Voronoi cell of robot $i$ \cite{rus}. Lloyd's algorithm reaches a local optima when all the robots are at the mass center of their respective Voronoi cells.

\subsection{Heterogeneous Robot Team Composition}
To solve the issue of the ground robots' limited sensing range, we propose the use of a heterogeneous team of robots consisting of $ K $ unmanned aerial robots and $ N $ unmanned ground robots. The ground robots operate in a domain $ \mathcal{D}^G \in \mathbb{R}^2 $ and the aerial robots operate in a domain $ \mathcal{D}^A \in \mathbb{R}^2 $, above and parallel to  $ \mathcal{D}^G $, surveying both the density field $ \phi (q) $ and the relative position of the ground robots below. In the proposed algorithm, the aerial robots can communicate with each other and the ground robots and exchange small amounts of data, including their global position, the relative ground robot locations, and cell weighting information to be defined later. The ground robots can receive the aforementioned data from the aerial robots and locally sense the domain within their sensing range.


\subsection{Combining Coarse Global Information with Local Sensing}

\begin{figure}
    \centering
    \includegraphics[width=\linewidth]{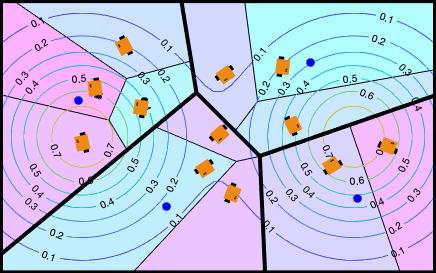}
    \caption{Visual demonstration of the relationships between the aerial Voronoi boundaries (thick borders), the ground Voronoi boundaries (colored patches), ground robot locations (orange cartoon figure) and aerial robot locations (blue dots) }
    \label{fig:voronoi}
\end{figure}

To adequately observe the regions of interest within domain, $\mathcal{D}^G$, for the ground robot team, the aerial robots perform standard Lloyd's algorithm over, $\mathcal{D}^A$, assuming a uniform distribution $\phi^A(q)$. When doing this, the aerial robots create regions of dominance in the form of Voronoi cells $ \mathcal{V}^A_j \subset \mathcal{D}^A, j = 1\dots K$. Each aerial robot will have access to two key pieces of information: number of robots $ n_j \in \mathcal{V}^A_j$ when $ \mathcal{V}^A_j $ is projected onto $ \mathcal{D}^G $ and the density function $ \phi^G(q) $ contained in $ \mathcal{V}^A_j $ projected on $ \mathcal{D}^G$. Figure \ref{fig:voronoi} demonstrates the relative relationships between the two sets of robots.
The aerial robots sense larger regions of the ground domain than the ground robots but at a coarse resolution; this coarse but general sensing allows the aerial robots to determine whether the coarse distribution of robots is adequate to cover the underlying importance function within their region of dominance, $\mathcal{V}^A_j$. The aerial robots may pass this information to drive the global distribution of the ground robots who may then use their finer, local sensing abilities to cover and survey the region with higher fidelity. This section develops the distributed ground robot controller that leverages the short range sensor information and coarse global information to effectively cover the domain.

With the number of robots and mass contained in a projected aerial cell, as well as the total number of ground robots deployed, $N$, we can formulate a measure that determines whether a region on the ground domain has too many or few ground robots. The goal is to align the number of ground robots contained in an aerial Voronoi projection with the amount of field density in an aerial Voronoi projection. Assuming $ \phi^G (q) $ is a probability distribution, the percentage of robots needed to cover the area contained in aerial Voronoi cell $j$ is,
\begin{equation} \label{PhiEqn}
 \Phi_j =   \frac{\int_{\mathcal{V}^A_j} \phi^G(q) \,dq}{\int_{\mathcal{D}^G} \phi^G(q) \,dq}.
\end{equation}

 Given the percentage of total density in an aerial Voronoi cell $\mathcal{V}^A_j$, $\Phi_j $, and $ N $ total ground robots deployed, the ideal number of robots within $ \mathcal{V}^A_j $, $n_{j,\text{ideal}}$, should be,
 
 

\begin{equation} \label{idealDistributionEqn}
     n_{j, \text{ideal}} = \Phi_jN .
\end{equation}


\noindent Using the relation in \eqref{idealDistributionEqn}, we can define an aerial Voronoi cell's underabundance or overabundance of robots by defining an aerial cell weight,


\begin{equation} \label{aerialCellWeightEqn}
    \sigma_j = \frac{n_j}{N} - \Phi_j.
\end{equation}

\noindent For \eqref{aerialCellWeightEqn}, an aerial cell with an insufficient amount of ground robots will have a negative weight and cells with a surplus of ground robots will have a positive weight. The cell weight is bounded between $[-1,1]$. 

Using $ \sigma_j $ as a measure to compare aerial cells, we can now formulate a control law that distributes robots to optimal regions over the domain $ \mathcal{D}^G $. We consider a control law that fuses traditional Lloyd's algorithm locally and robot distribution globally. As stated earlier, Lloyd's algorithm implements the following control law;
\begin{equation} \label{localLaw}
     u_{i,\text{local}} = \kappa (c(p_i) - p_i),
\end{equation}

where $ \kappa > 0$ is a scalar controller gain, $p_i \subset \mathcal{D}^G \in \mathbb{R}^2$ is the location of robot $i$, and $ c(p_i) $, which depends on density function $\phi^G(q)$ and $p_i$, is the mass center of Voronoi cell of robot $ i $ over domain $ \mathcal{D}^G $.

We define a global distribution control law $u_{i, \text{global}}$ with the following,
\begin{equation} \label{globalLaw}
     u_{i, \text{global}} = \gamma ( C_{\text{min}} - p_i),
\end{equation}
where $\gamma >0$ is a controller gain and $C_\text{min}$, which depends on locations of the aerial robots and $\sigma_j$, is the geometric center of the aerial Voronoi cell with the minimum cell weight $\sigma_\text{min}$,
\begin{equation} \label{sigMin}
     \sigma_\text{min} = \min\limits_j \sigma_j.
\end{equation}

To combine the local and global controllers, we define the control law in equation (\ref{heteroLaw}), 

\begin{equation} \label{heteroLaw}
     u_i = (1 - \hat{\sigma}_j) u_{i,\text{local}} + \hat{\sigma}_j u_{i, \text{global}},
\end{equation}
to continuosly switch between local coverge and global distribution depending on a given robot's necessity in an aerial Voronoi cell. The weighting variable $\hat{\sigma}_j$,
\[ \hat{\sigma}_j = \begin{cases} 
     \frac{n_j}{N} - \int_{\mathcal{V}^A_j} \phi^G(q) \,dq  & \frac{n_j}{N} - \int_{\mathcal{V}^A_j} \phi(q) \,dq  > 0 \\
      0 & \frac{n_j}{N} - \int_{\mathcal{V}^A_j} \phi(q) \,dq  \leq 0 \\
   \end{cases},
\]
balances the aerial robots' global coverage information with the ground robots' local coverage information to dynamically favor global distribution or local coverage based on the number of robots within each aerial Voronoi cell. The goal of this control law is to leverage the coarse global information of the aerial robots to overcome the limited sensing of the ground robots' high resolution local coverage. In order to use $ {\sigma}_j $ in the control law, $\hat{\sigma}_j$ is defined as $ {\sigma}_j $ bounded above $0$. This is due to the fact that allowing $\hat{\sigma}_j<0$ could possibly result in robots settling to an undesired local minima on the boundary between deprived aerial cells (i.e. with $ {\sigma}_j < 0$).


\subsection{Controller Stability}

According to \cite{cortes}, Lloyd's algorithm finds a stable and locally optimal solution to coverage. Thus, to prove stability, we show that each cell weight $\sigma_j$ approaches $0$ as $t\rightarrow \infty$, reducing the proposed controller in equation \eqref{heteroLaw} to the controller in equation \eqref{localLaw}, which is proven asymptotically stable in \cite{cortes}.

In this proof, each grond robots' influence is approximated as a bivariate normal distribution over $\mathcal{D}^G$ with mean $p_i$ and variance $\Sigma$. The number of robots in an aerial cell $j$ is approximated by integrating all robots' distributions over the aerial Voronoi cell $j$. This creates a continuum approximation of $\frac{n_j}{N}$.

\begin{equation} \label{continuum}
    \frac{n_j}{N} \approx \frac{1}{N} \sum_{i=1}^N  \int_{\mathcal{V}^A_j} \mathcal{ N }(p_i, \Sigma) dq
\end{equation}

\begin{equation} \label{sigContinuum}
    \sigma_j \approx   \frac{1}{N} \sum_{i=1}^N  \int_{\mathcal{V}^A_j} \mathcal{ N }(p_i, \Sigma) dq -  \int_{\mathcal{V}^A_j} \phi^G (q) dq
\end{equation}

\noindent With the assumption that the field density $\phi^G(q)$ is normalized and the number of robots in the system is static, the total sum of the aerial cell weights $\sigma_j$ is always equal to zero,

\begin{equation} \label{sumTot}
     \sum_{j=1}^K  \sigma_j = 0.
\end{equation}

\noindent Due to the conservation of robots assumption in \eqref{sumTot}, the minimum cell weight $\sigma_\text{min}$ will always be less than or equal to zero when not in equilibrium. Intuitively, this states that if there are aerial cells with too many robots, there must also be aerial cells with too few robots. In this proof, we show that the minimum cell weight is always approaching zero and thus the total magnitude of all cell weights is also approaching zero due to the conservation relationship described in equation (\ref{sumTot}). 


The minimum cell weight is always increasing because the control law defined in equation (\ref{heteroLaw}) actively attracts ground robots to the aerial cell with the lowest weight, by definition, and drives robots out of aerial cells with too many robots. To prove this, we will show the time derivative of $\sigma_\text{min}$ is always positive when the system is under the control law defined in equation (\ref{heteroLaw});

$$ \frac{d\sigma_\text{min}}{dt} = \sum_{i=1}^N  \langle \nabla \sigma_{\text{min},i} , u_i \rangle$$
 
\noindent where $\nabla \sigma_{\text{min},i}$ is the gradient of $\sigma_\text{min}$ due to robot $i$ and $u_i$ is the control produced by equation \eqref{heteroLaw}. 

Due to the proportional nature of the global and local controllers, if $\sigma_j \neq 0$, the control law $u_i$ will be dominated by $u_{i,\text{global}}$, $u_i \approx u_{i,\text{global}}$. Thus, the time derivative of the minimum cell weight can be calculated as follows;


$$ \frac{d\sigma_\text{min}}{dt} = \sum\limits_{i=1}^N \langle \nabla \sigma_\text{min} , u_{i,\text{global}} \rangle. $$ 
When taking the gradient of $\sigma$ with respect to the location of the robot, the field density term $\int_{\mathcal{V}^A_\text{min}} \phi (q) dq$ goes to zero because neither the aerial Voronoi cell nor the density function within an aerial Voronoi cell depend on the ground robots' locations and we are left with the gradient of the integral of the normal distribution. In order to perform this definite Gaussian integral, we can approximate the bounds of the aerial Voronoi cell by defining a bounding box, $\mathcal{B}$, that completely contains the aerial cell.

$$ \sigma_{\text{min}} = \int_{\mathcal{V^A_{\text{min}}}}\mathcal{N}(p, \Sigma) dq \leq \int_{\mathcal{B}}\mathcal{N}(p, \Sigma) dq, $$

$$ \nabla \sigma_\text{min} \leq \frac{1}{N} \int_\mathcal{B}  \nabla \mathcal{ N }(p, \Sigma) dq, $$

\begin{equation} \label{grad}
\nabla \sigma_\text{min} \leq \alpha (e^{(b + p)^2} -e^{(b - p)^2} )
\end{equation}

\noindent where $b$ is the bound of the box around the Voronoi cell of interest, $\alpha$ is a positive scalar that is an artifact of integrating a normal distribution that does not change the direction of gradient, and $p$ refers to a robot's location with respect to the geometric center of the bounding box. 


With the gradient of the lowest weight, $\sigma_\text{min}$, defined in \eqref{grad}, the lowest weight is always increasing if the inner product between $\nabla \sigma_\text{min}$ and the global control law $u_{i,\text{global}}$ is positive. Since $u_{i,\text{global}}$ is, by definition, pointed towards the center aerial cell with the lowest weight and the gradient of the robot's normal distribution with respect to its position is pointed inwards towards the cell with the lowest weight (as shown in \ref{grad}), we can conclude that their inner product is positive.

\begin{equation}
    \frac{d\sigma_\text{min}}{dt} = \sum\limits_{i=1}^N \langle \nabla \sigma_\text{min} , u_{i,\text{global}} \rangle > 0
\end{equation}
$$ \lim_{t\to\infty} \sigma_{\text{min}} = 0$$
 
 It is important to note that the cell that has the lowest weight can change with time; this is because as robots move into a cell that is in need of robots, its weight increases and thus no longer needs robots so another aerial cell will become the lowest weighted cell. With the relationship described in equation (\ref{sumTot}), we know that the magnitude of all the weights will decrease as $t\rightarrow \infty$. As the magnitude of all the weights go to zero, the control law defined in equation (\ref{heteroLaw}) approaches the standard Lloyd's algorithm control law, which has been proven stable in \cite{cortes}. 
 
 \begin{figure}[t]
    \centering
    \includegraphics[width=\linewidth]{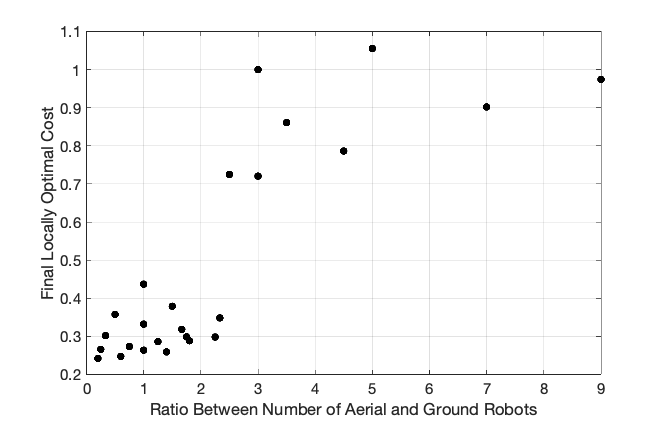}
    \caption{Computed final cost of robot teams surveying a bimodal distribution with differing values of $\frac{K}{N}$ }
    \label{fig:reasonablesness}
\end{figure}

\begin{figure*}[!t]
    \centering
    \begin{subfigure}{0.33\textwidth}
    \includegraphics[trim={1.85cm 2.2cm 1.85cm 2.2cm}, clip, width=\linewidth]{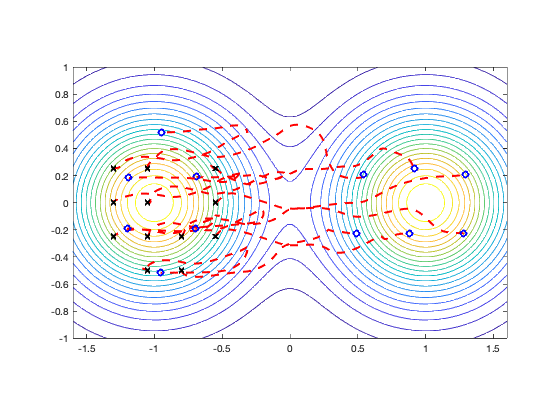}%
    \caption{}
    \label{fig:hetero_lloyd}%
    \end{subfigure}\hfill
    \centering
    \begin{subfigure}{0.33\textwidth}
    \includegraphics[trim={1.85cm 2.2cm 1.85cm 2.2cm}, clip, width=\linewidth]{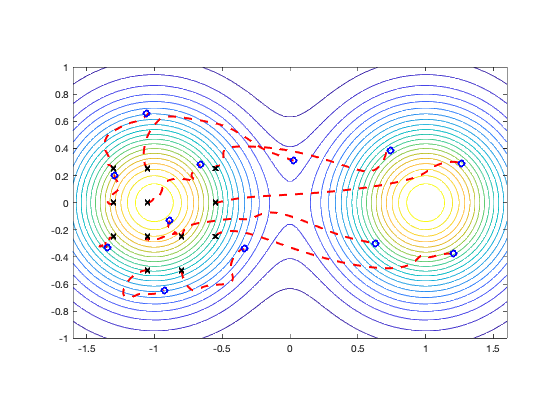}%
    \caption{}
    \label{fig:homo_lloyd}%
    \end{subfigure}\hfill
    \centering
    \begin{subfigure}{0.33\textwidth}
    \includegraphics[trim={1.85cm 2.2cm 1.85cm 2.2cm}, clip, width=\linewidth]{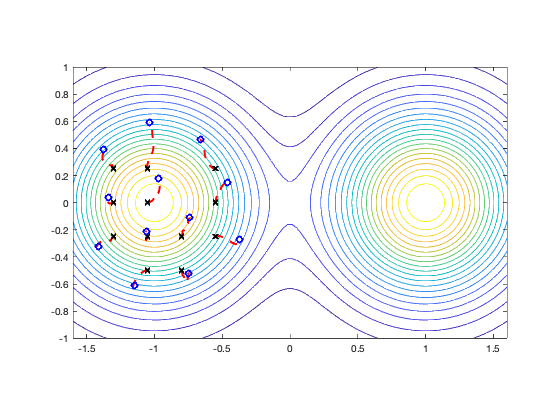}%
    \caption{}
    \label{fig:sensing_radius}%
    \end{subfigure}\hfill
    \centering
    \begin{subfigure}{0.33\textwidth}
    \includegraphics[trim={0.5cm 0.5cm 0.5cm 0.5cm}, clip, width=\linewidth]{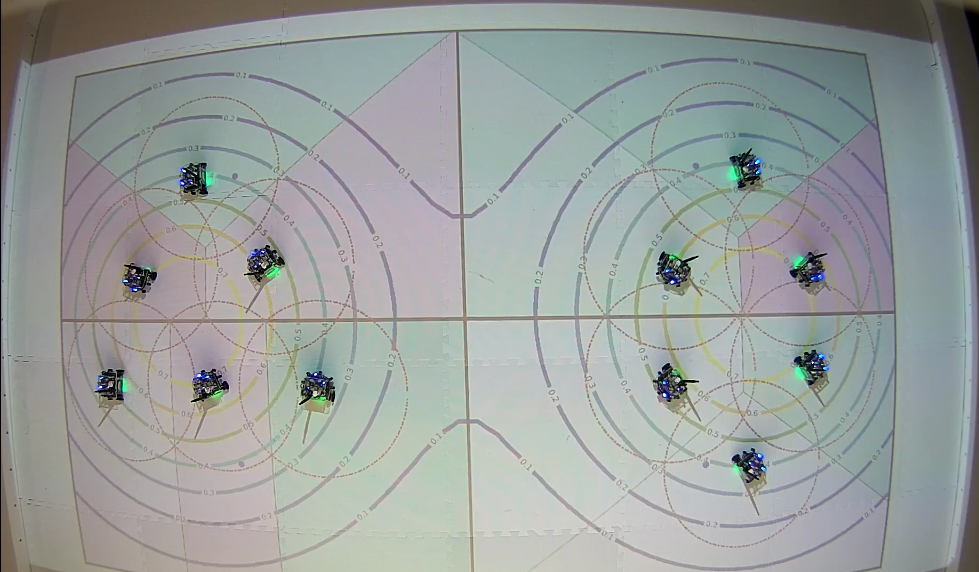}%
    \caption{}
    \label{fig:hetero_lloyd_robot}%
    \end{subfigure}\hfill
    \centering
    \begin{subfigure}{0.33\textwidth}
    \includegraphics[trim={0.5cm 0.5cm 0.5cm 0.5cm}, clip, width=\linewidth]{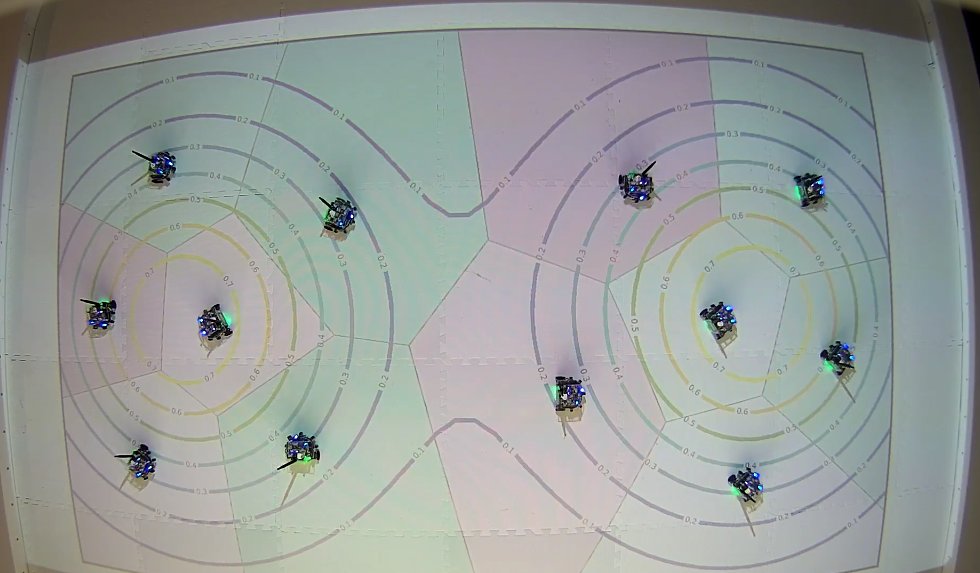}%
    \caption{}
    \label{fig:homo_lloyd_robot}%
    \end{subfigure}\hfill
    \centering
    \begin{subfigure}{0.33\textwidth}
    \includegraphics[trim={0.5cm 0.5cm 0.5cm 0.5cm}, clip, width=\linewidth]{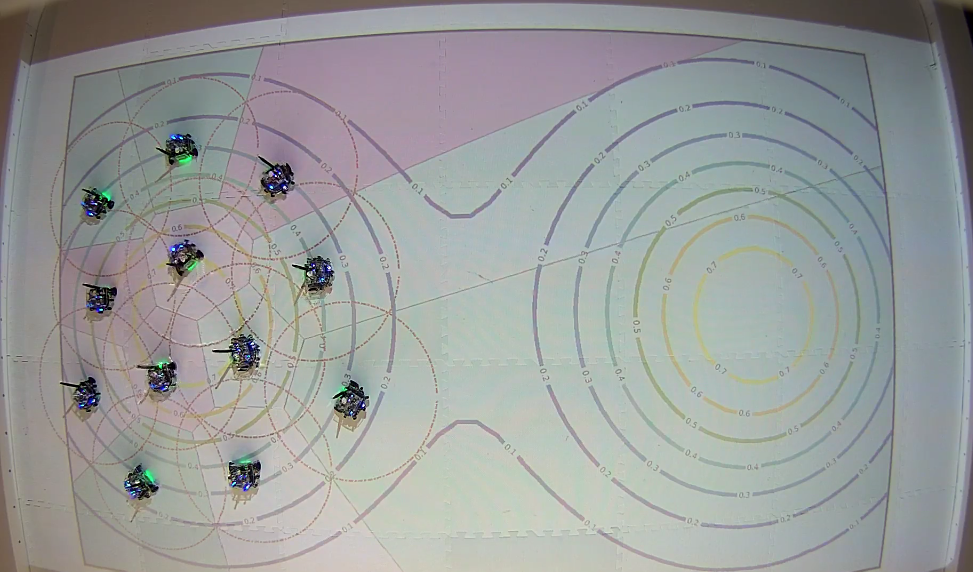}%
    \caption{}
    \label{fig:limited_lloyd_robot}%
    \end{subfigure}%
    \caption{State trajectory of the 12 ground robots in simulation (a) - (c) and the final configuration in experiment (d) - (f) using, (a),(d) the proposed algorithm, (b),(e) standard Lloyd's algorithm with unlimited sensor range, and (c),(f) standard Lloyd's with limited sensing range. For the simulation figures, we indicate the initial positions of the ground robots (black X markers), the trajectory of each robot (red dotted line), and the final position of the ground robots (open blue circles).}
    \label{fig:results}
\end{figure*}

\section{Experimental Results}
In this section, we will present simulated and experimental results of the proposed algorithm and compare to previous solutions to coverage control. To validate the algorithm performance, the \textit{Robotarium} \cite{robotarium}, a remotely accessible, multi-robot research facility at the Georgia Institute of Technology is used. We will also present necessary modifications to the proposed algorithm that improve performance due to the discrete nature of the system (i.e. robots do not have continuous influence).

\subsection{Varying the Ratio of Robots}
First, it is important to note that the number of aerial and ground robots in the system can vastly change the algorithm's performance. When the ratio of aerial robots to ground robots, $\frac{K}{N}$, is large, the ground robots will be directed towards only the peaks of the non-uniform field over $\mathcal{D}^G$ and artifacts of the due to the discrete positioning of the robots will adversely influence the behaviour of the algorithm will become more apparent. When this ratio is kept small, the ground robots can find an appropriate balance between distributing themselves in their respective cells and exploring other cells. However, as more ground robots are added, the communication bandwidth requirement for the aerial robots increases due to the higher information transmission demand to pass information to the ground robots. The trivial case would be to set $K=1$ and only vary $N$ however, this disregards the ground robots' range limited nature. Thus, an appropriate ratio can be chosen given the ground robots sensing range and domain size. Figure \ref{fig:reasonablesness} shows the steady state final value of the normalized cost functions of various $\frac{K}{N}$ ratios of robots operating over the same domain $\mathcal{D}^G$ with the same underlying density field. To evaluate the quality of coverage, we use the cost function defined in equation (\ref{cost}) evaluated at each time iteration of the system. For Figure \ref{fig:reasonablesness} we are using the cost from equation (\ref{cost}) evaluated once the stationary local optimum is reached.

\subsection{Consideration of Discrete Robot Positions}
From the definition in equation (\ref{aerialCellWeightEqn}) we know that, unless $N$ is infinite, $\sigma_j$ is going to take on discrete values. Thus, when a single ground robot enters or leaves an aerial cell, that cell's weight will change by $\pm \frac{1}{N}$. This is an issue because the definition of the ideal distribution of robots defined in equation (\ref{idealDistributionEqn}) can never be achieved. To avoid Zeno effects \cite{zeno}, the discretization error of $\sigma_j$ should be bounded to $\pm \frac{1}{N}$. In the implementation of \eqref{heteroLaw} the following modification is made to $\hat{\sigma}_j$,
\begin{equation}
 \hat{\sigma}_j = \begin{cases} 
     \frac{n_j}{N} - \int_{\mathcal{V}^A_j} \phi^G(q) \,dq , & \frac{n_j}{N} - \int_{\mathcal{V}^A_j} \phi^G(q) \,dq  > \frac{1}{N} \\
      0, & \frac{n_j}{N} - \int_{\mathcal{V}^A_j} \phi^G(q) \,dq  \leq \frac{1}{N} \\
   \end{cases}.
\end{equation}
This modification prevents robots from oscillating between cells with $|\sigma_j| < \frac{1}{N}$.


\subsection{Experimental Results}
The proposed algorithm is implemented on the \textit{Robotarium} using simulated aerial robots and differential drive robots. Barrier certificates are implemented on the testbed to guarantee safe and collision free operation. The \textit{Robotarium} allows robots to operate on a 320cm by 200cm rectangular domain.

For this experiment, we choose the underlying distribution of the ground domain $\mathcal{D}^G$ to be a bivariate Gaussian distribution. This multi-modal distribution is chosen because, depending on the initial positions of the ground robots, a range-limited team might settle to an undesirable local minimum and not be able to effectively observe both modes of the domain.



To show the relative performance of this proposed algorithm, we present the coverage capabilities on the bimodal distribution of three robot teams each consisting of $N=12$ ground robots. The first team performs standard Lloyd's algorithm with no modifications (i.e. assuming unlimited range sensing) assuming the aforementioned bimodal density. The second team also performs standard Lloyd's algorithm but consists of robots with range-limited sensors with a maximum sensing distance of $30$ centimeters. The final team performs the proposed algorithm with range-limited sensors (also with a maximum sensing distance of $30$ centimeters) and $K=4$ aerial robots where the aerial robots first perform standard Lloyd's algorithm with an uniform distribution.
\begin{figure}
  \includegraphics[width=\linewidth]{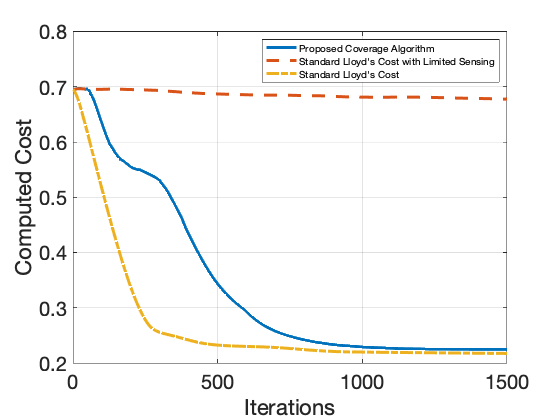}
  \caption{Cost $\mathcal{H}$ for each of the three simulated scenarios.}
  \label{fig:simcost}
\end{figure}
\begin{figure}
  \includegraphics[width=\linewidth]{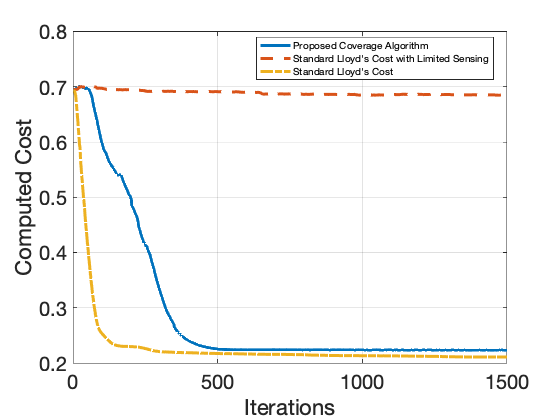}
  \caption{Cost $\mathcal{H}$ for each of the three experimental scenarios.}
  \label{fig:expcost}
\end{figure}
Figures \ref{fig:hetero_lloyd}, \ref{fig:homo_lloyd}, and \ref{fig:sensing_radius} demonstrate the state trajectories of each of the three scenarios in simulation. Additionally, the three scenarios were tested identically to their simulated analog on \textit{Robotarium} testbed using physical robots and the final configurations are shown in Figures \ref{fig:hetero_lloyd_robot}, \ref{fig:homo_lloyd_robot}, and \ref{fig:limited_lloyd_robot}. It is not guaranteed that the proposed algorithm approaches the same configuration as standard Lloyd's algorithm. However, from Figure \ref{fig:simcost} and \ref{fig:expcost} we see that a range-limited team, performing this paper's proposed algorithm, achieves a similar quality of coverage to that of a range-unlimited team performing standard Lloyd's algorithm. With restricted information, the team performing range-limited coverage control on this domain settles in a local optimum of the coverage cost function that under performs relative to the range-limited team performing the proposed algorithm of this paper.

\section{Conclusion}

In this paper, we proposed a method to use a heterogeneous team of robots to perform coverage control on a domain with an unknown density function. We leveraged aerial robots with long-range, coarse resolution sensors to define general regions of interest on the domain and use ground robots with fine, short-range sensors to locally cover the general regions of interest. To make use of these two distinct robot teams, we formulated a novel extension to Lloyd's algorithm that fuses the global distribution information from the aerial robots and the local coverage information from the ground robots. Experiments and simulation showcasing the capabilities of this proposed algorithm were performed to show its performance compared with standard methods. We have proposed a method to perform coverage control using a heterogeneous team of aerial and ground robots that leverages the distinct sensing capabilities of each robot type.

\bibliographystyle{IEEEtran}
\bibliography{main}

\begin{thebibliography}{10}
\providecommand{\url}[1]{#1}
\csname url@samestyle\endcsname
\providecommand{\newblock}{\relax}
\providecommand{\bibinfo}[2]{#2}
\providecommand{\BIBentrySTDinterwordspacing}{\spaceskip=0pt\relax}
\providecommand{\BIBentryALTinterwordstretchfactor}{4}
\providecommand{\BIBentryALTinterwordspacing}{\spaceskip=\fontdimen2\font plus
\BIBentryALTinterwordstretchfactor\fontdimen3\font minus
  \fontdimen4\font\relax}
\providecommand{\BIBforeignlanguage}[2]{{%
\expandafter\ifx\csname l@#1\endcsname\relax
\typeout{** WARNING: IEEEtran.bst: No hyphenation pattern has been}%
\typeout{** loaded for the language `#1'. Using the pattern for}%
\typeout{** the default language instead.}%
\else
\language=\csname l@#1\endcsname
\fi
#2}}
\providecommand{\BIBdecl}{\relax}
\BIBdecl

\bibitem{guatam}
A.~{Gautam} and S.~{Mohan}, ``A review of research in multi-robot systems,'' in
  \emph{2012 IEEE 7th International Conference on Industrial and Information
  Systems (ICIIS)}, 2012, pp. 1--5.

\bibitem{cao}
Y.~U. {Cao}, A.~S. {Fukunaga}, A.~B. {Kahng}, and F.~{Meng}, ``Cooperative
  mobile robotics: antecedents and directions,'' in \emph{Proceedings 1995
  IEEE/RSJ International Conference on Intelligent Robots and Systems. Human
  Robot Interaction and Cooperative Robots}, vol.~1, 1995, pp. 226--234 vol.1.

\bibitem{ota}
J.~Ota, ``Multi-agent robot systems as distributed autonomous systems,''
  \emph{Advanced Engineering Informatics}, vol.~20, no.~1, pp. 59 -- 70, 2006.

\bibitem{Pimenta2}
L.~C.~A. Pimenta, M.~Schwager, Q.~Lindsey, V.~Kumar, D.~Rus, R.~C. Mesquita,
  and G.~A.~S. Pereira, \emph{Simultaneous Coverage and Tracking (SCAT) of
  Moving Targets with Robot Networks}.\hskip 1em plus 0.5em minus 0.4em\relax
  Berlin, Heidelberg: Springer Berlin Heidelberg, 2010, pp. 85--99.

\bibitem{lowenberg}
J.~{Lowenberg-DeBoer}, I.~{Huang}, V.~{Grigoriadis}, and S.~{Blackmore},
  ``Economics of robots and automation in field crop production,''
  \emph{Precision Agriculture}, vol.~21, pp. 278--299, 2020.

\bibitem{zhong}
M.~{Zhong} and C.~G. {Cassandras}, ``Distributed coverage control and data
  collection with mobile sensor networks,'' \emph{IEEE Transactions on
  Automatic Control}, vol.~56, no.~10, pp. 2445--2455, 2011.

\bibitem{cortes}
J.~{Cortes}, S.~{Martinez}, T.~{Karatas}, and F.~{Bullo}, ``Coverage control
  for mobile sensing networks,'' \emph{IEEE Transactions on Robotics and
  Automation}, vol.~20, no.~2, pp. 243--255, 2004.

\bibitem{arslan}
O.~{Arslan} and D.~E. {Koditschek}, ``Voronoi-based coverage control of
  heterogeneous disk-shaped robots,'' in \emph{2016 IEEE International
  Conference on Robotics and Automation (ICRA)}, 2016, pp. 4259--4266.

\bibitem{hexsel}
B.~{Hexsel}, N.~{Chakraborty}, and K.~{Sycara}, ``Coverage control for mobile
  anisotropic sensor networks,'' in \emph{2011 IEEE International Conference on
  Robotics and Automation}, 2011, pp. 2878--2885.

\bibitem{cortesRangeLimited}
J.~Cort\'{e}s, S.~Martin\'{e}z, and F.~Bullo, ``Spatially-distributed coverage
  optimization and control with limited-range interactions,'' \emph{ESAIM:
  Control, Optimisation and Calculus of Variations}, vol.~11, pp. 691--719,
  October 2005.

\bibitem{santos}
M.~{Santos}, Y.~{Diaz-Mercado}, and M.~{Egerstedt}, ``Coverage control for
  multirobot teams with heterogeneous sensing capabilities,'' \emph{IEEE
  Robotics and Automation Letters}, vol.~3, no.~2, pp. 919--925, 2018.

\bibitem{rus2}
A.~Breitenmoser, M.~Schwager, J.-C. Metzger, R.~Siegwart, and D.~Rus, ``Voronoi
  coverage of non-convex environments with a group of networked robots,''
  \emph{IEEE International Conference on Robotics and Automation}, 2010.

\bibitem{kantaros}
Y.~Kantaros, M.~Thanou, and A.~Tzes, ``Distributed coverage control for concave
  areas by a heterogeneous robot–swarm with visibility sensing constraints,''
  \emph{Automatica}, vol.~53, pp. 195 -- 207, 2015.

\bibitem{cortesVariations}
J.~{Cortes}, S.~{Martinez}, T.~{Karatas}, and F.~{Bullo}, ``Coverage control
  for mobile sensing networks: Variations on a theme,'' \emph{Mediterranean
  Conferenceon Control and Automation}, 2002.

\bibitem{pimenta}
L.~C.~A. {Pimenta}, V.~{Kumar}, R.~C. {Mesquita}, and G.~A.~S. {Pereira},
  ``Sensing and coverage for a network of heterogeneous robots,'' in \emph{2008
  47th IEEE Conference on Decision and Control}, 2008, pp. 3947--3952.

\bibitem{robotarium}
S.~Wilson, P.~Glotfelter, L.~Wang, S.~Mayya, G.~Notomista, M.~Mote, and
  M.~Egerstedt, ``The robotarium: Globally impactful opportunities, challenges,
  and lessons learned in remote-access, distributed control of multirobot
  systems,'' \emph{IEEE Control Systems Magazine}, vol.~40, no.~1, pp. 26--44,
  2020.

\bibitem{rus}
M.~Schwager, D.~Rus, and J.-J. Slotine, ``Unifying geometric, probabilistic,
  and potential field approaches to multi-robot deployment,'' \emph{The
  International Journal of Robotics Research}, vol.~30, no.~3, pp. 371--383,
  2011.

\bibitem{zeno}
M.~{Heymann}, {Feng Lin}, G.~{Meyer}, and S.~{Resmerita}, ``Analysis of zeno
  behaviors in a class of hybrid systems,'' \emph{IEEE Transactions on
  Automatic Control}, vol.~50, no.~3, pp. 376--383, 2005.

\end{thebibliography}

\end{document}